\documentclass[conference]{IEEEtran}
\IEEEoverridecommandlockouts
\usepackage{cite}
\usepackage{amsmath,amssymb,amsfonts}
\usepackage{algorithmic}
\usepackage{graphicx}
\usepackage{textcomp}
\usepackage{xcolor}
\usepackage{tikz,hyperref}
\usepackage{url}
\usepackage[T1]{fontenc}
\def\BibTeX{{\rm B\kern-.05em{\sc i\kern-.025em b}\kern-.08em
    T\kern-.1667em\lower.7ex\hbox{E}\kern-.125emX}}
    
\definecolor{lime}{HTML}{A6CE39}
\DeclareRobustCommand{\orcidicon}{
	\begin{tikzpicture}
	\draw[lime, fill=lime] (0,0) 
	circle [radius=0.16] 
	node[white] {{\fontfamily{qag}\selectfont \tiny ID}};
	\draw[white, fill=white] (-0.0625,0.095) 
	circle [radius=0.007];
	\end{tikzpicture}
	\hspace{-2mm}
}    
\foreach \x in {A, ..., Z}{\expandafter\xdef\csname orcid\x\endcsname{\noexpand\href{https://orcid.org/\csname orcidauthor\x\endcsname}
			{\noexpand\orcidicon}}
}

\newcommand{\AlgoName}{COSTI}

\begin{document}

\title{COSTI: a New Classifier for Sequences \\of Temporal Intervals
\thanks{The~project was funded by POB Research Centre for Artificial Intelligence and Robotics of Warsaw University of Technology within the~Excellence Initiative Program -- Research University (ID-UB).}
}

\author{\IEEEauthorblockN{Jakub Micha\l ~Bilski}
\IEEEauthorblockA{\textit{Faculty of~Mathematics and Information Science} \\
\textit{Warsaw University of~Technology}\\
Warsaw, Poland \\
jakub.michal.bilski@gmail.com}
\and
\IEEEauthorblockN{Agnieszka~Jastrz\k{e}bska}
\IEEEauthorblockA{\textit{Faculty of~Mathematics and Information Science} \\
\textit{Warsaw University of~Technology}\\
Warsaw, Poland \\
A.Jastrzebska@mini.pw.edu.pl
}
}

\maketitle

\begin{abstract}
Classification of sequences of temporal intervals is a part of time series analysis which concerns series of events. We propose a~new method of transforming the problem to a task of multivariate series classification. We use one of the state-of-the-art algorithms from the latter domain on the new representation to obtain significantly better accuracy than the state-of-the-art methods from the former field. We discuss limitations of this workflow and address them by developing a~novel method  for classification termed COSTI (short for Classification of Sequences of Temporal Intervals) operating directly on sequences of temporal intervals. The proposed method remains at a high level of accuracy and obtains better performance while avoiding shortcomings connected to operating on transformed data. We propose a~generalized version of the problem of classification of temporal intervals, where each event is supplemented with information about its intensity. We also provide two new data sets where this information is of substantial value.
\end{abstract}


	\section{Introduction}
	
    This paper is concerned with classifying sequences of temporal intervals, a type of data that consists of sequences of different intervals limited by start and end timestamps. 
    
    The contribution of this paper is threefold. Firstly, we introduce a methodology for the generalization of classification of sequences of temporal intervals by adding information about the event's intensity. This generalization widens the scope of practical applications of algorithms from this domain. We share two new data sets concerning music analysis and meteorology where intensity values are of substantial, informative importance. The proposed scheme can be applied to a plethora of real-world domains. Example novel areas of application include:
    \begin{itemize}
    \item healthcare (one sequence corresponds to one patient, events correspond to symptoms or reactions to drugs, each symptom/reaction can be described with its intensity that can vary in time), 
    \item business process mining (one sequence corresponds to one process, intensity depicts the number of people participating in a given stage/event in the process),
    \item smart homes (one sequence corresponds to one house, intensity values can be used to describe number of persons in a room, energy consumption, etc.).
    \end{itemize}
    The list above is not exclusive. The new approach opens the room for novel studies on temporal events analysis where it is essential to measure the time of a~change with high accuracy.
    
    Secondly, we show how to convert data available in a~generalized representation format to multivariate time series, a~widely known time-varying phenomena representation form. This transformation allows using algorithms developed for the latter data representation format to be applied to the former. We show that using MiniRocketMultivariate \cite{minirocket}, one of the multivariate time series classification methods, on the converted data gives considerably better results than dedicated state-of-the-art methods in the classification of sequences of temporal intervals (SMILE \cite{smile} and Stife \cite{stife}). However, we can only apply the described transformation between data representation formats after establishing a sampling step, which is a~challenging problem in itself. Moreover, utilization of sampling introduces time complexity depending on the duration of the sequences divided by the sampling step. When high accuracy is needed, and the duration of sequences is very high, this leads to a very high execution time. Based on the observed facts, we postulate that there is a need for an algorithm that would work directly on interval data (without sampling), not depend on the magnitude of timestamps and perform at a high level of accuracy.

    The culminating aspect of the contribution is a new algorithm termed COSTI (short for Classification of Sequences of Temporal Intervals) for generalized classification of sequences of temporal intervals, adapting the core ideas of the MiniRocketMultivariate algorithm in a new method that works very well on a~different type of~input data, with some necessary conceptual changes and a~completely new implementation. While retaining a high level of accuracy, the presented algorithm does not depend on the duration of sequences in time complexity but rather on the number of intervals present in the data. As a~result, the new algorithm is faster than MiniRocketMultivariate for each considered real-world dataset; it is over 50 times faster in two cases.  We show that the new algorithm is well-fitted for sequences of temporal intervals.
    Not only is the new method applicable to the generalized version of the problem, but it also performs on a level of accuracy higher than state-of-the-art methods when the constrained version of the problem is concerned.

	An indispensable part of the study presented here is the implementation of the algorithm. It was written in Python and compiled via Numba \cite{numba}. Our code was made available at \url{https://bit.ly//3sTiHup}. In the paper, we address an empirical study concerning eight datasets, which are at the moment deemed as benchmark datasets in this domain \cite{smile}.
	
	The remainder of the paper is structured as follows. Section~\ref{sec:Literature} introduces preliminary notions and relevant literature positions. Section \ref{sec:ClassificationThroughTransformation} demonstrates the proposed data format transformation procedure and discusses its limitations. Section \ref{sec:Method} describes the proposed new method of classification of sequences of temporal intervals. Section \ref{sec:data_sets} introduces two new data sets and identifies assumptions made in the experiments part. Section \ref{sec:Experiments} examines results of the conducted experiments. Section \ref{sec:Conclusion} is devoted to conclusions and critical discussion.
	
	\section{Existing solutions}
	\label{sec:Literature}
	\subsection{Problem statement: multivariate time series classification -- even steps in~time}
	A common assumption in data analysis of time-varying phenomena is that the data are present as a sequence of real numbers, and successive values were collected with some fixed, evenly spaced interval. We were not given an analytic form of the relation describing the phenomenon in time. Still, we imagine it as some continuous real-valued function that we can approximate by analyzing data points collected in a time series. Technically, we can represent such data as a~vector containing real numbers indexed with a timestamp. We can say that this vector includes data from a single channel of information. In a situation where a~given time-varying phenomenon is observed using more than one channel, we can store such data in a~matrix, where successive columns correspond to consecutive information channels, and a~timestamp is used to index the rows. An important assumption is to maintain the same sampling rate in each channel.

	\subsection{Multivariate time series classification -- even steps in~time}
	
	As Pasos Ruiz et al.~\cite{Pasos2021} suggest, Dynamic Time Warping (DTW)-based approaches are deemed to be  the `gold standard' solution to time series classification. It is often used as a bottom-line method with which more sophisticated approaches are compared. DTW, in its simple form, is suitable for univariate time series. However, the literature proposes a couple of generalizations that allow to apply it in a scheme aiming at multivariate time series classification.  As Shokoohi-Yekta et al.~\cite{Shokoohi2017} outline, the most straightforward  adaptation is to run the procedure independently for all variables in the time series and then combine the results. The second strategy assumes that warping is the same across all dimensions \cite{Zhao2020}. Some approaches adaptively incorporate independent and dependent warping strategies  \cite{Shokoohi2015}.
	
	When it comes to algorithms dedicated to multivariate data, we shall mention Multiple Representation Sequence Learner, MrSEQL \cite{Nguyen2019} which is an ensemble operating on symbolic time series representation and deep neural architectures addressed by Fawaz et al.~\cite{Fawaz2020}.
		
	Finally, we shall mention that univariate time series classifiers can be adapted to deal with multivariate time series. We can list several studies addressing this line of research, where for example, HIVE-COTE  \cite{Middlehurst2021} or RISE \cite{Lines2018} algorithms were applied in a~new role. The problems arise less in terms of classification quality but in classifiers' performance. In some cases, including the case of a~classifier called WEASEL \cite{Schaefer2018} or CIF \cite{Middlehurst2020}, one can find several ways to enhance the algorithm's efficiency to process multivariate data faster.

	\subsection{Univariate time series classification using Rocket}
	
	Rocket \cite{rocket}, introduced in~2019, achieved a~state-of-the-art accuracy while requiring only a~fraction of~the training time required by the~existing methods of~univariate time series classification. Rocket uses random convolutional kernels to~create a~set of~features. Convolutional kernel is a~concept that was first utilized in~convolutional neural networks. A~convolutional kernel $f$ as used in~Rocket can be expressed as (version with no padding):
	
	\begin{equation}
	    f(X) = \bigg\{ (\sum^{L-1}_{k=0} w_k \cdot X[t+dk]) - b \bigg\}_{t=1,...,T-(L-1)d}
	\end{equation}
	where $X \in R^T$ is the~input vector, $L \in N$ is the~length of~the kernel, $w_k \in R$ is its k-th weight, $b \in R$ is bias and $d \in N$ is dilation. When padding is enabled, a~number of~zeros is appended to~$X$ at its beginning and end, so that the~iteration of~$t$ can start with the~middle weight getting multiplied by $X[1]$ and end by the~middle weight getting multiplied by $X[T]$. Please note that $f: R^T \rightarrow R^{T-dL}$ or $f: R^T \rightarrow R^{T}$, depending on whether padding is used, so one convolutional kernel applied to~one sequence produces many real numbers.
	
	Rocket generates a~high number of~random kernels, i.e., kernels with random length, weights, bias, dilation, and padding. After a~kernel is applied on a~time series, two statistics are calculated based on the~output -- maximum value and proportion of~positive values. These two statistics become new features describing the~sequence. Normally, a~very high number of~kernels is used (10$\,$000), generating a~set of~20$\,$000 features for each sequence. Then, a~classifier is used to~perform the~final classification. Most often, Ridge Classifier is used, sometimes replaced by a~single softmax layer when number of~examples is very big.
	
	\subsection{Univariate time series classification -- even steps in~time: MiniRocket}
	
	Even though Rocket was extremely fast comparing to~other algorithms in~the~field, its authors were able to~come up with a~version up to~75 times faster on larger data sets. MiniRocket \cite{minirocket}, introduced in~2020, was able to~retain the~same level of~accuracy as Rocket, while greatly reducing the~time of~execution and making the~whole process almost deterministic (or fully deterministic, at an additional computational expense). This was done mainly by reducing to~a~bare minimum the~random space from which the~convolutional kernels were generated. Length of~the kernel in~MiniRocket is fixed to~9 and all~random kernels must contain three weights equal to~$-2$ and six weights equal to~1, creating a~set of~84 possible kernels. Maximal value statistic is discarded, making proportion of positive values the only method used for feature generation. Possible dilation values now consist of~32 integer values evenly distributed over an exponential distribution over interval $[1, T/8]$. Also, the choice of~biases now relies on creating a~histogram of~a convolutional output for a~random training example, from which random quantiles are extracted and become biases. Many other small improvements were also introduced. These all~changes made it possible to~introduce many computational shortcuts contributing to~the~performance of~the method. Authors declare that from now on, MiniRocket should be used as a~default version of~the algorithm.

	\subsection{Multivariate time series classification -- even steps in~time: MiniRocketMultivariate}
	Although MiniRocket was developed as an algorithm for univariate time series, a~basic extension for multivariate case was also created by the~same authors. The~basic premise of~this algorithm is treating data in~the~same way as in~MiniRocket, only now instead of~multiplying the~weight by one element, it gets multiplied by sum of~elements from a~subset of~channels. This subset is chosen at random individually for each feature.

	\subsection{Problem statement: classification of~sequences of~temporal intervals}
	\label{sec:SecondTypeOfData}
    The subsequent generalization considered in this paper is to move to the case where the classification task concerns a sequence of temporal events with assigned duration. Each event is assigned a type, a start time, and an end time. The end time is always greater than or equal to the start time. When it is equal, according to Kotsifakos et al.~\cite{ibsm}, we say that an event is instantaneous. We assume that no events of the same type can overlap, while events of different types can.
    
    It is possible to extend this description by adding a weight to an event, representing its intensity. Additionally, we can describe the absence of an event as an event with a  weight value equal to zero. With this encoding, it is possible to represent this type of phenomena in a manner analogous to the case of multivariate time series--the proposed generalization results in an event being interpreted as an information channel.

	\subsection{Classification of~sequences of~temporal intervals}
	\label{sec:SmileParts}
	Classification of~sequences of~temporal intervals has received substantially less attention than the~main stream of~time series classification. To~our best knowledge, there~are only two algorithms that constitute the~current state of~the art. Those are Stife and SMILE, the~latter being an improved version of~the former. SMILE works by creating a~vector of~features for each sample. Extracted features are then used to~perform classification using Random Forest \cite{Probst2019}. Features are generated in~four ways:
	\begin{enumerate}
	    \item \textbf{Static features}: a~set of~14 predefined statistics. These range from very simple, like total number of~events in~a~sequence, to~more complex, like summed duration of~time intervals when there was a~maximal number of~events active at the~same time.
	    \item \textbf{Class-based medoid distance features}: distances to~medoids representing each class. Medoids are chosen from training data based on IBSM, a~distance metric for sequences of~temporal intervals, introduced by Kotsifakos et al.~\cite{ibsm}.
	    \item \textbf{Interval relation-pair features}: numbers of~occurrences of~seven Allen's temporal relations, introduced by Allen~\cite{atr}. These relations can occur between pairs of~events and the~number of~their occurrences is calculated for each configuration of~different types of~events (e.g.~relation ``X precedes Y'' between event of~type 2 and event of~type 4 occurs 19 times in~the~sequence, 19 is the~feature). Only 75~features from this group are chosen to~be included in~the~final set of~features. This choice is made based on information gain calculated on the~training set.
	    \item \textbf{Interval segment features}: a~set of~distances between the~sequence and 75~subsequences randomly extracted from the~training set. The~function used to~measure distance is called ABIDE \cite{abide}.
	\end{enumerate}
	Features generated by these four methods are concatenated and passed to~a~Random Forest classifier. Workflow of~Stife is very similar, only it does not use Interval segment features.

	\section{Classification through transformation between data representation formats}
	\label{sec:ClassificationThroughTransformation}

	The introduced data transformation method opens the possibility to apply methods of classification operating on the latter type of data, which can be very beneficial given a high development of the field of multivariate series classification. In addition, we show that using MiniRocketMultivariate on the transformed data gives considerably better accuracy than the current state-of-the-art  described in Section~\ref{sec:SmileParts}.
	
	\subsection{Transforming sequences of~temporal intervals into multivariate time series}
    \label{sec:TransformingProblems}

	\begin{figure}
	    \centering
	    \includegraphics[scale=0.155]{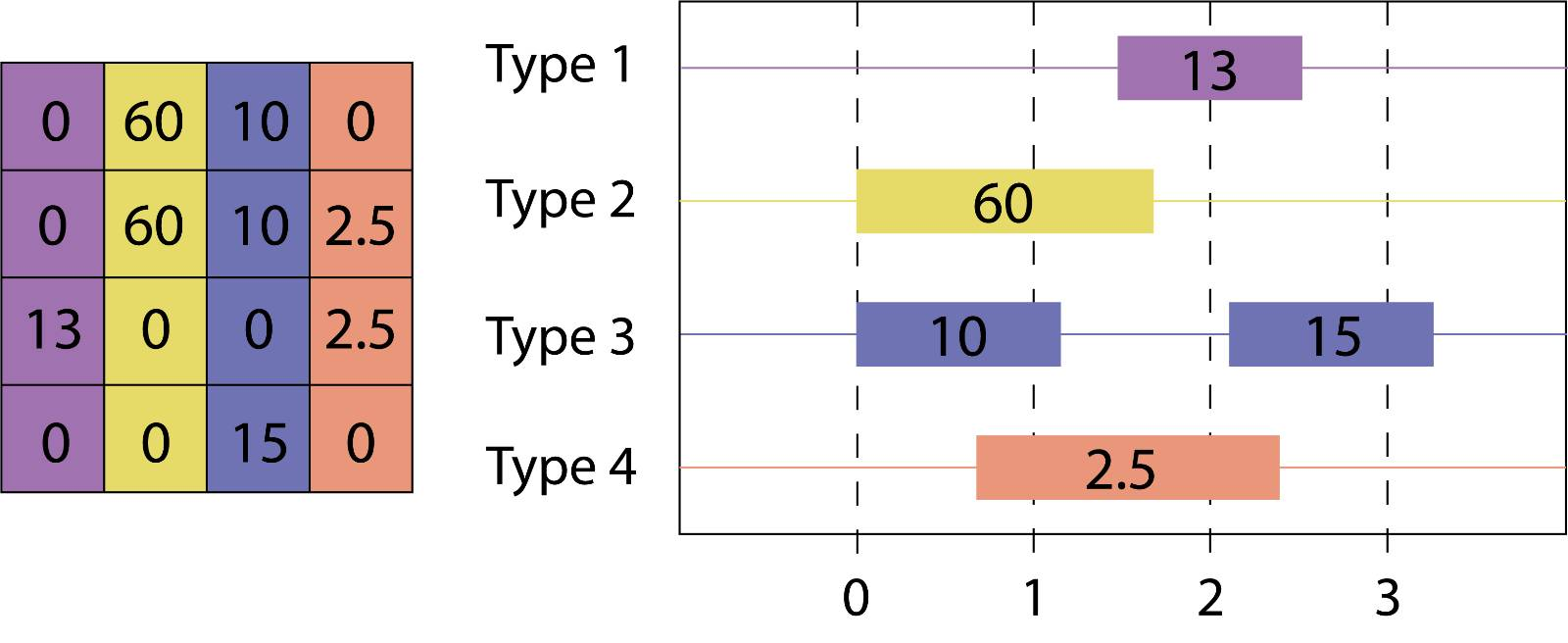}
	    \caption{Left: data structure suitable to store multivariate time series. Right: example sequence of events considered in a problem of sequences of temporal intervals classification.}
	    \label{fig:representations}
	\end{figure}
	
    Figure~\ref{fig:representations} illustrates an example of multivariate time series data and an event sequence to indicate how conceptually transparent the transition between one format and the other is. The general idea is to sample channels (events) at a~chosen frequency. We can only do this after choosing a step in time. Using that step, consecutive rows in~the~matrix can be populated with values of~channels, according to~the~formula: $M[r,c] = X_c[r \cdot s]$, where $r$ is row, $c$ is column and $s$ is the~chosen step. In a particular case of MiniRocketMultivariate, we want all matrices to be of the same height, as the algorithm operates on sequences of equal length. However, when it is implicit that time in~$X$ starts from 0, it is not always explicitly stated what the~end of~the sequences is. In that case, we can only depend on the~last timestamp of~a sequence, i.e., the~end of~the last interval. Thus, the~number of~rows of~the matrix is set to~the~number of~steps resulting from parsing the~sequence that contains the~highest end timestamp. All~other sequences are effectively padded with zeros. Other techniques of ensuring the same size of input matrices are also possible and we describe one such technique in Section \ref{sec:MiniRocketMultivariateComparison}.
    
    A number of difficulties can arise as a consequence of operating on the data created with the described transformation. Too high a sampling rate unnecessarily increases the volume of data. Too low a frequency, in turn, results in too much information being lost. Selecting the right frequency is difficult and demands domain knowledge from the user. Sometimes even the best rate will give unsatisfactory results because the first variant is not suitable for storing the second type of information efficiently. These not-yet-solved issues motivated us to develop a new method for classification of sequences of temporal intervals, operating directly on the interval data. We compare the performance of our solution with the results obtained using transformation between formats and show that not only it allows to omit the problematic sampling stage, it also benefits from operating on the original data in terms of time efficiency.

	\section{Method}
	\label{sec:Method}
	
	In this section, we present our new approach to the task of sequences of temporal intervals classification. The~new algorithm can be conceptually summarized as a~processing pipeline consisting of~the following steps:
	\begin{enumerate}
	    \item Transforming interval-based data set to~a~custom data structure.
	    \item Fitting model parameters using a~training set.
	    \item Calculating features for training and test sets.
	    \item Performing data classification using Ridge Classifier.
	\end{enumerate}
    The proposed solution can be seen as a generalization of the time series processing steps present in MiniRocketMultivariate algorithm.

	\subsection{Introductory notes on the new approach}
	\label{sec:MethodPreliminaries}
	Calculation of a single feature is performed using a set of randomly chosen parameters: a subset of channels $C$, bias $b$, dilation $d$ and a convolutional kernel with weights $\{w_0,…,w_8 \}$. Now, let us consider a case when observations $X$ are defined on a continuous range $<t_s,t_e>$. This would allow us to introduce the following way of computing features:

	\begin{equation}
	\label{eqn:continuous_feature}
	    a~= \frac{1}{t_e-t_s} \int_{t_s}^{t_e} H \bigg( (\sum^8_{k=0} \sum_{c \in C} w_k \cdot X_c[t+dk])-b \bigg) dt
	\end{equation}
	where $H(\cdot)$ is the~Heaviside step function given as:
	
	\begin{equation}
	    H(x) = \left\{
	    \begin{array}{ll}
        1, &\textrm{ $x>0$}\\
        0, &\textrm{ $x \leq 0$}\\
        \end{array} \right.
	\end{equation}  
	
	Please note that the~value of~the integral is effectively the~sum of~lengths of~all intervals where the~expression inside $H(\cdot)$ is larger than zero.
	
	From now on, let us assume that each channel of~$X$ is defined by a~number of~intervals of~constant values, as described in~Section \ref{sec:SecondTypeOfData}. We can utilize this characteristic to~create an efficient method of~computing the~integral in~(\ref{eqn:continuous_feature}). The~key approach here will be aligning sequences corresponding to~each k in~$X_c[t+dk]$, and then creating a~sorted concatenation using those. This will allow us to~calculate the~integral by just iterating over a~sequence of~timestamps between which the~value of~the expression inside $H( \cdot )$ does not change. Below the~algorithm description, you may find a~toy example illustrating this approach.

	\subsection{The new algorithm -- steps}
	\label{sec:AlgorithmDescription}
	\begin{enumerate}
	    \item Create a list $L_0$ containing all timestamps where the value of all channel changes is sorted according to the timestamp in ascending order. Store information about the difference between the new and the old value and the channel's id where the change happened. 
	    \item $L = L_0 \bigcup L_1 \bigcup ... \bigcup L_8$, where $L_k$ is created by adding $dk$ to~each element of~$L_0$. Save the~information about $k$ which the~element originated from. This information will be used to~choose the~weight of~the convolutional kernel.
	    \item Sort all~elements of~$L$ by their timestamps in~ascending order. Please note that the~individual lists $L_k$ are already sorted, which can be used to~greatly speed up the~procedure.
	    \item Set $psum=-b$, $a=0$.
	    \item Traverse $L$, adding to~$psum$ the~changes of~values of~intervals multiplied by the~weight corresponding to~the~$k$ from which the~change point originated from. Take into account only changes in~the~channels belonging to~the~subset $C$. Whenever $psum$ over the~current interval is larger than zero, add the~length of~the interval to~$a$.
	    \item Feature = $a/(t_e-t_s)$.
	\end{enumerate}
	Please note that the~first three points of~this algorithm demand only the~value of~the dilation to~be known. As the~original MiniRocketMultivariate algorithm uses a~maximum of~32 different dilation values, we suggest to analogously compute the~lists in~points 1-3 for these 32 dilations, and for each feature calculate only points 4-6.

	\subsection{Illustrative Example}

	\begin{figure}
	    \centering
	    \includegraphics[scale=0.16]{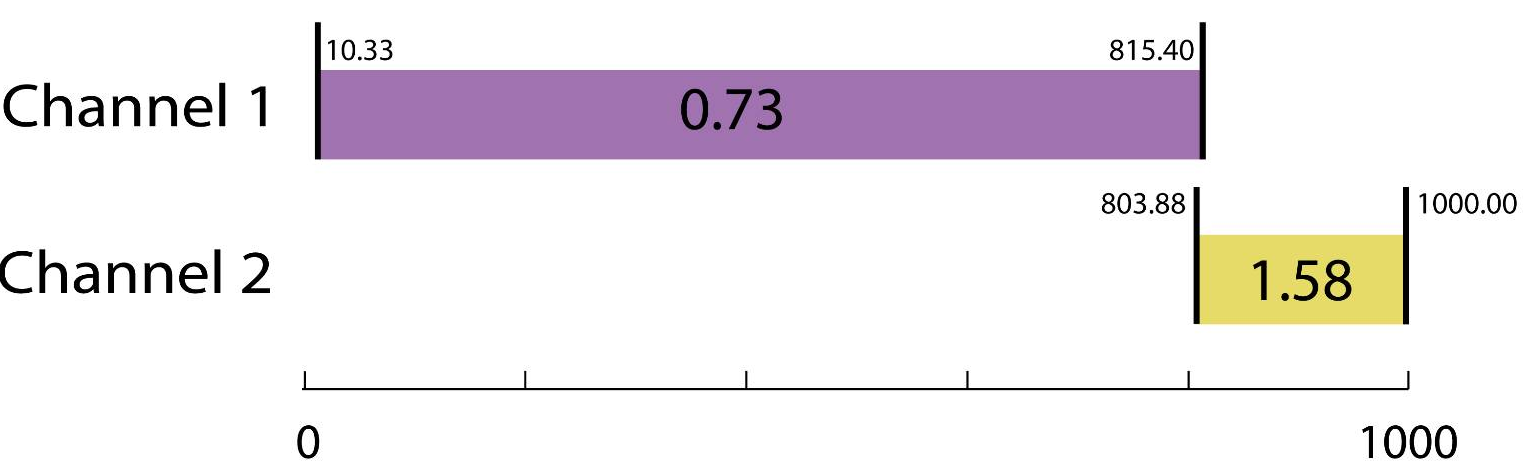}
	    \caption{Example sequence consisting of two events.}
	    \label{fig:toy_example}
	\end{figure}

    \begin{figure}
	\begin{center}
    \begin{tabular}{ |c|c|c|c| } 
     \hline
     Timestamp & Value & Channel & Weight index \\ 
     \hline
     \hline
     10.33 & 0.73 & 1 & 1 \\ 
     \hline
     30.33 & 0.73 & 1 & 2 \\
     \hline
     803.88 & 1.58 & 2 & 1 \\
     \hline
     815.40 & -0.73 & 1 & 1 \\
     \hline
     823.88 & 1.58 & 2 & 2 \\
     \hline
     835.40 & -0.73 & 1 & 2 \\
     \hline
     1000.00 & -1.58 & 2 & 1 \\
     \hline
     1020.00 & -1.58 & 2 & 2 \\
     \hline
    \end{tabular}
    \end{center}
    \caption{Toy case study data structure containing $L$.}
    \label{fig:toy_example_table}
    \end{figure}

    Fig.~\ref{fig:toy_example_table} presents $L$, a data structure created based on the sequence visualized in Fig.~\ref{fig:toy_example}. Please note that the consecutive rows in the $L$ data structure correspond to changes in value of the expression \textit{psum}~$=\sum^8_{k=0} \sum_{c \in C} w^c_k \cdot X_c[t+dk]-b$, only in this toy example 8 is replaced by 1.
    This can be visualized as a convolutional kernel of length equal to
    2 moving above the sequence along the time axis. When one of its inputs approaches the next change point, a new value of the expression is computed by adding $w_k \cdot v$ to psum, where $v$ is ``value'' and $k$ is ``weight index''. Before we update \textit{psum}, we check if the previous value was greater than 0. If it was, the $H(\cdot)$ function would return 1, so we can add the length of the interval that just ended to the final integral value $a$.

	Data structure created this way can be used to~calculate the~integral for any given kernel and bias. Moreover, if we want to~take a~smaller combination, say $C={1}$, we can just ignore timestamps that have ``channel'' field equal to~$2$ during the~structure iteration.
	
    \subsection{Selecting final timestamp}
    We denote the highest timestamp found in the data by $t_{max}$ and set $t_{min}$ always equal to 0. These values are then used to determine $t_s$ and $t_e$, depending on the usage of padding (see Section \ref{sec:Padding}). This approach can be seen as padding all sequences with events with intensity equal to zero to meet duration of the longest sequence.
	
	\subsection{Fitting biases}
	Choice of~bias depends on a~histogram of $\sum^8_{k=0} \sum_{c \in C} w^c_k \cdot X_c[t+dk]$, from which random quantiles are extracted and become biases. Here, $X$ is an element from the training set, chosen at random for each combination of convolutional kernels and dilation values. Extraction of~a random quantile can be performed using a~cumulative sum of~duration over intervals of~occurring values sorted by values. This technique does not create any serious computational overhead.
	
	\subsection{Fitting dilations}
	The fact that the number of used dilations is limited to 32~directly contributes to achieving high performance using this workflow. The new algorithm can use floating-point numbers for dilation values as observations are no longer defined only in fixed timestamps. Please note that MiniRocketMultivariate dilation distribution starts at 1, as no smaller dilation can be used. No such lower boundary can be established in the presented solution, as timestamps can be of any magnitude. Thus, dilations are taken from evenly divided exponential distribution over interval $[d_{min}, (t_{max}-t_{min})/8]$, where $d_{min}$ is the smallest difference between different timestamps found in the training data. Using continuous values means also avoiding duplicates (in MiniRocketMultivariate there can be multiple equal dilations, if $T$ is small enough). This could, in~theory, convey to~a~higher modelling power. However, experiments on data containing only integer timestamps show that the solution performs better when dilations are rounded to the nearest integers. This may suggest that when there is a~strong sampling frequency present in the data, it is beneficial to use only multiples of this frequency as dilations. However, the experimental data was too limited to decide on the significance of this relationship. Also, we did not want to permit the usage of floating-point timestamps. As a compromise, dilation values are rounded to integers only when it is detected that all timestamps in train data are integers.
	
	\subsection{Padding}
	\label{sec:Padding}
	Similarly to MiniRocketMultivariate, we use padding in half of the features. The way features are computed when padding is on corresponds to~$t_s=t_{min}+4d$, $t_e=t_{max}+4d$, so the~interval depends on~the~middle weight remaining in~the~range of~the original data (taking into consideration positive shifts in dilation introduced in (\ref{eqn:continuous_feature})). The~case with no padding is $t_s=t_{min}+8d$, $t_e=t_{max}$. No padding is used when biases are generated.
	
	\subsection{Time complexity}
	
	Let us address time complexity of the proposed approach and compare it with MiniRocketMultivariate, which was the starting point for our research. 
	
	MiniRocketMultivariate has linear scalability in the number of features ($m$), number of examples ($n$) and time series length ($l_{input}$). The number of channels chosen for computation of a~single feature is taken from a logarithmic distribution, which makes the overall complexity equal to $O(m \cdot n  \cdot l_{input} \cdot log(C))$, where $C$ is the number of channels. When used to classify sequences of intervals, MiniRocketMultivariate takes a set of matrices as an input, which normally must be generated from the interval data. The complexity of this operation is proportional to the number of cells in all matrices, which is $O(n \cdot l_{input} \cdot C)$.

	Creating the data structure and calculating features in COSTI comes with time complexity $O(m \cdot n \cdot e \cdot log(C))$, where $e$~is the number of all events present in the data. Additionally, sorting the input sequences $L_0$ at the beginning is $O(n \cdot e \cdot log(e))$. Please note that step 3~of the algorithm described in Section~\ref{sec:AlgorithmDescription} involves concatenating and sorting 9~sorted lists. This is implemented as iterating over 9~sequences using a priority queue, achieving linear time complexity.
	
	\begin{center}
    \begin{tabular}{ |c|c| } 
     \hline\hline
     
     \multicolumn{2}{|c|}{MiniRocketMultivariate} \\
     \hline
     
     \multicolumn{2}{|c|}{$O(m \cdot n  \cdot l_{input} \cdot log(C) + n \cdot l_{input} \cdot C)$} \\
     \hline
     \hline
     \multicolumn{2}{|c|}{\AlgoName} \\
     \hline
     \multicolumn{2}{|c|}{
     $O(m \cdot n \cdot e \cdot log(C) + n \cdot e \cdot log(e))$}\\
     \hline
     \hline
     \multicolumn{2}{|c|}{$m$ - number of~features} \\
     \multicolumn{2}{|c|}{$n$ - number of~samples} \\
     \multicolumn{2}{|c|}{$e$ - number of~events} \\
     \multicolumn{2}{|c|}{$C$ - number of~channels} \\
     \multicolumn{2}{|c|}{$l_{input}$ - number of~rows in~a~matrix} \\
     \hline
    \end{tabular}
    \end{center}

	\section{Data sets and basic experiment assumptions}
	\label{sec:data_sets}
	The proposed methodology of transitions and generalizations between time series data sets and sequences data sets is a novel contribution of this paper. Furthermore, the literature does not offer any solutions designed to~solve the problem of~classification of~sequences of~temporal intervals after generalization we outlined in~Section \ref{sec:SecondTypeOfData} (by adding values corresponding to~events). 
	
	In Section \ref{sec:musekey} and Section~\ref{sec:weather}, we present two new data sets that we contribute alongside this paper. They concern real-world phenomena that are best described as sequences of events with meaningful weights (which is an extended temporal data representation format, more versatile than sequences of events with no weights). 
	
	\subsection{Existing data sets}
	Nonetheless, to~compare the~new solution with other algorithms, we also had to~perform tests on a~constrained version of~the classification of~event intervals problem, with no values, just events happening or not. There exists a~small benchmark data collection for this problem, containing six data sets. Recent algorithms in~this field, Stife \cite{stife} and SMILE \cite{smile}, used this data for accuracy evaluation.

	Please note that we did not prepare a specialized version of the algorithm despite using the algorithm for a constrained problem, where timestamps are integers and there are no values associated with intervals.
	For the use of comparison with other algorithms, all values were set to 1 when an event occurs (and to 0 when it does not, but this is also true for an unconstrained problem). This was also true for the two new data sets.
	
	\subsection{New dataset for musical key recognition}
	\label{sec:musekey}
	The data set contains time series extracted from 240 MIDI files taken from the MASETRO Dataset \cite{maestrodataset} and from the public domain. The~files contained recordings of piano pieces, either created manually or recorded during live performance. We processed each MIDI file to~create a~series of~events spanning across 12 channels. Each channel corresponds to~one of~the notes (C, C\#$=$Db, D, D\#$=$Eb, E, F, F\#$=$Gb, G, G\#$=$Ab, A, A\#$=$Bb, B). An event in~a~channel is present when at least one key corresponding to~the~channel's note is being pressed. Timestamps in~the~time series were taken from the~MIDI files, as these also operate on timestamps. However, timestamps in~MIDI are purely relative -- they cannot be easily converted to~time, and it is impossible to~compare them across different files. Considering this, we decided to~resize all~timestamps, so the~end of~the last interval in~each recording would be equal to~24000. This value should be big enough to~ensure that all~rounding errors are negligible (for a~ten minutes-long piece, this is 40 Hz, and in~a~normal tempo like \textit{moderato}, one thirty-second note lasts 1/12 of~a second). In a version of~the data set with values associated with intervals, value of~an interval is set to~the~velocity the~key was pressed with. The~classification task is to~predict which of~the 24 key signatures the~piece was written in. There are 12 classes with 20 samples each and 12 classes with ten samples each.

	\subsection{New dataset concerning weather events}
	\label{sec:weather}
	The data set was extracted from a~weather events data set, US Weather Events (2016 - 2020), a~countrywide data set of~6.3 million weather events. We pulled only the~events recorded in~2017 by weather stations in~the~state of~Louisiana. Each channel corresponds to~the presence of~some specific type of~weather conditions. Individual series were made from the~recordings for every~single weather station and each single month, summing up to~12 series for a~single station. Each series contains a~month of~weather events, with timestamps rounded with 10 minutes step, counting from the~month's start (please note, that even using a~one second step would not hinder the~performance of~the presented algorithm. However, this would be hardly reasonable given the~nature of~the data).
	In a~version of~the data set with values associated with intervals, the value of~an interval is equal to~the intensity of the weather condition. The classification task is to predict which month the series was recorded in. For a~detailed description of~the data and levels of intensity, please refer to~\cite{weather}.
	
	\begin{table}
	\begin{center}
    \caption{Summary of~the eight data sets used in~experimentation. Musekey and Weather are the new datasets contributed with this paper.}
    \begin{tabular}{ |c|c|c|c|c|c|c| } 
     \hline
     Data set & Classes & Samples & Channels & Max duration & Events\\
     \hline
     Auslan2 & 10 & 200 & 12 & 30 & 12.24 \\ 
     \hline
     Blocks  & 8 & 210 & 8 & 123 & 5.75 \\
     \hline
     Context & 5 & 240 & 54 & 284 & 80.65 \\
     \hline
     Hepatitis & 2 & 498 & 63 & 7555 & 108.28 \\
     \hline
     Pioneer & 3 & 160 & 92 & 80 & 55.93 \\
     \hline
     Skating & 6 & 530 & 41 & 6829 & 43.78 \\
     \hline
     Musekey & 24 & 360 & 12 & 24000 & 2914.61 \\
     \hline
     Weather & 12 & 540 & 7 & 4465 & 63.16 \\
     \hline
    \end{tabular}
    \end{center}
    \label{tab:caption}
	\end{table}
	
	\section{Results}
	\label{sec:Experiments}
	\subsection{Comparison with state-of-the-art}
	There are two state-of-the-art algorithms in~the~field of~classification of~sequences of~temporal intervals: Stife and SMILE. No implementation of~SMILE was made available to~the~public up to~the~time this paper was submitted, so we followed \cite{smile} and recreated SMILE using python and Numba. 
	We decided to~use our own implementation of~Stife, even though its implementation, written in~Java, is available in~the~Internet. 
	Both hyperparameters of~SMILE, that is number of~\textit{2-lets} and number of~\textit{e-lets}, were set to~75, as suggested in~the~paper. 
	We used 75 as the~number of~\textit{2-lets} in~Stife. As for the accuracy comparison, we followed the~example of~\cite{smile} and included two extracted parts of~SMILE, namely Static features (STAT) and Class-based medoid distance features (MEDOID), described in~Section \ref{sec:SmileParts}. These were used as individual methods of~feature extraction based on which classification was performed.
	Default version of~a Random Forest classifier from scikit-learn package \cite{scikit-learn} was used as the~final step in~all~four above methods. Additionally, we included IBSM method, which SMILE was also compared to. IBSM method works by \mbox{1-nearest-neighbour} classification using the~IBSM distance. As for the~presented algorithm, we used default values of~10 000 features and 32 dilations. All~experiments were performed using 10-fold cross-validation. Every experiment was repeated 10 times and mean values of~accuracy were calculated.

	\begin{table}
	\begin{center}
    \caption{Accuracy comparison with state-of-the-art.}
    \label{tab:SotaComparison}
    \begin{tabular}{ |c|c|c|c|c|c|c|} 
    \hline
    Data set & STAT & MEDOID & IBSM & STIFE & SMILE & \AlgoName \\
    \hline
    Auslan2 & 0.331 & 0.437 & 0.415 & 0.503 & 0.509 & \textbf{0.579} \\
    Blocks & 0.840 & \textbf{1.000} & \textbf{1.000} & \textbf{1.000} & \textbf{1.000} & \textbf{1.000} \\
    Context & 0.947 & 0.961 & 0.967 & 0.991 & 0.991 & \textbf{0.996} \\
    Hepatitis & 0.661 & 0.697 & 0.603 & 0.799 & 0.795 & \textbf{0.841} \\
    Pioneer & 0.822 & 0.823 & 0.949 & 0.977 & 0.975 & \textbf{1.000} \\
    Skating & 0.599 & 0.902 & 0.969 & 0.954 & 0.970 & \textbf{0.994} \\
    Musekey & 0.143 & 0.280 & 0.039 & 0.473 & 0.473 & \textbf{0.768} \\
    Weather & 0.611 & 0.495 & 0.603 & 0.685 & 0.680 & \textbf{0.909} \\
    \hline
    \end{tabular}
    \end{center}
    \end{table}

    Results of experiments are presented in Table \ref{tab:SotaComparison}. The new method achieved a significant improvement over the current state-of-the-art in the field of classification of sequences of temporal intervals, reaching the highest scores for all eight data sets.
    
	\subsection{Comparison with MiniRocketMultivariate}
    \label{sec:MiniRocketMultivariateComparison}
    In~all~the~experiments, step during matrices creation $s=1$, so we simulate a~situation when no information is discarded. Matrices creation for sequences that have different highest timestamps can be performed by padding shorter sequences with zeros, as explained in Section \ref{sec:TransformingProblems}. An alternative approach would be to~perform bilinear interpolation on each matrix that has a~number of~rows smaller than maximal, stretching it to meet the number of rows of the longest matrix. A similar technique is used in IBSM distance \cite{ibsm} when two sequences of different length are compared. Decision on the matrix creation technique could be made with regard to~the~domain knowledge. We decided to~test both approaches on each data set. Please note that this dilemma concerns only MiniRocketMultivariate. \AlgoName operates directly on the~interval data and does not use matrix representation.

	In all tests, we used the~implementation of~MiniRocketMultivariate available through the~sktime package. In all algorithms, default numbers of~features (10~000) and dilations (32) were used. Time of~execution of~MiniRocketMultivariate did not include creating a~matrix representation of~the data. Time of~execution of~the new algorithm did include sorting the~data at the~beginning. All~experiments were performed using \mbox{10-fold} cross-validation. Every experiment was repeated 10 times and mean values of~accuracy and time were calculated.
	
	\begin{table}
	\begin{center}
    \caption{Comparison with MiniRocketMultivariate used on a~matrix representation.}
    \label{tab:MiniRocketMultivariate}
    \begin{tabular}{ |c|c|c|c| } 
    \hline
    & \multicolumn{3}{c|}{accuracy} \\
    \hline
     & \multicolumn{2}{c|}{MiniRocketMultivariate} & \AlgoName \\
    \hline
    Data set & Scaling & Padding &\\
    \hline
    Auslan2 & 0.557 & 0.570 & \textbf{0.579}  \\
    Blocks & \textbf{1.000} & \textbf{1.000} & \textbf{1.000} \\
    Context & \textbf{0.996} & \textbf{0.996} & \textbf{0.996} \\
    Hepatitis & 0.829 & 0.824 & \textbf{0.841} \\
    Pioneer & \textbf{1.000} & \textbf{1.000} & \textbf{1.000} \\
    Skating & 0.987 & 0.992 & \textbf{0.994} \\
    Musekey & 0.755 & 0.752 & \textbf{0.768} \\
    Weather & 0.881 & \textbf{0.910} &  0.909 \\
    \hline
    & \multicolumn{3}{c|}{time [s]} \\
    \hline
    & \multicolumn{2}{c|}{MiniRocketMultivariate} & \AlgoName \\
    \hline
    Data set & Scaling & Padding &\\
    \hline
    Auslan2 & 0.59 & 0.59 & \textbf{0.52} \\
    Blocks & 1.00 & 1.08 & \textbf{0.55} \\
    Context & 2.18 & 2.25 & \textbf{1.18} \\
    Hepatitis & 134.44 & 133.39 & \textbf{2.80} \\
    Pioneer & 0.76 & 0.80 &  \textbf{0.67} \\
    Skating & 108.05 & 102.73 & \textbf{1.69} \\
    Musekey & 257.54 & 258.27 & \textbf{184.47} \\
    Weather & 31.04 & 31.55 & \textbf{5.57} \\
    \hline
    \end{tabular}
    \end{center}
    \end{table}
    
	Table \ref{tab:MiniRocketMultivariate} presents accuracy and time of~execution for all eight data sets. Although the presented solution achieved slightly better results in accuracy than the two variants of MiniRocketMultivariate, the difference is too little to consider it statistically significant. A much bigger difference in favour of the presented solution was observed for the time of execution. Please note that compared algorithms differ significantly in~terms of~important factors affecting the~time complexity. For example, the higher the maximal timestamp is, the longer it takes to execute both versions of MiniRocketMultivariate. On the other hand, dealing with a very high number of~events makes \AlgoName slower. Experimental results confirm this -- the~presented solution easily beats MiniRocketMultivariate in~efficiency on data sets with a~high maximal timestamp (Hepatitis, Skating) and performs at a similar level when the maximal timestamp is very low (Auslan2, Pioneer) or the number of events is very high (Musekey). This is not surprising. The~real conclusion resulting from this experiment is different: the~new algorithm was successfully implemented with a~level of~efficiency that makes using it on real data profitable when compared to~MiniRocketMultivariate with a~matrix representation. Not only faster, using it should also be easier, as it allows to use real values of~timestamps, and no domain knowledge to~establish a~sampling step is needed.
	
	\section{Conclusion and Critical Discussion}
	\label{sec:Conclusion}
	We believe that the main contribution presented in \AlgoName~does not come from its speed advantage over MiniRocketMultivariate on matrix representation, but rather from a new approach to processing information, more fitting to the problem of classification of interval sequences. Possibility of using real values of timestamps and avoiding any type of sampling is important not only when the modelling power is concerned. It can also prove to be advantageous from a practical perspective, making the framework easier to use for users with various levels of experience. We strongly believe there are many ways to improve accuracy of the presented method. One of possible research directions would be to examine the impact of different distributions of floating-point dilations. Other ways of analysing the convolution output sequences, more elaborate than the proportion of positive values, can also be considered.
	A weak spot of the proposed approach is its memory complexity. When we consider input data in a typical format (each interval is represented by three real values: start, end, and intensity, and one integer value describing the channel), the size of the data structure $L$ used in the algorithm is about $18$ times the size of the input. For the biggest data set we use (Musekey) this is roughly equal to 400 megabytes. When much bigger data sets emerge in the field, this trait of the method can become a significant downside, making it unsuitable for severely resource-constraint domains, like wearable electronics. However, we believe that mainstream domains of application of the developed approach are not falling into this area.

\bibliographystyle{IEEEtran}
\bibliography{mybibfile}
\vspace{12pt}

\end{document}